\documentclass[10pt,twocolumn,letterpaper]{article}

\usepackage[pagenumbers]{cvpr}

\definecolor{cvprblue}{rgb}{0.21,0.49,0.74}
\usepackage[pagebackref,breaklinks,colorlinks,allcolors=cvprblue]{hyperref}

\usepackage{amsfonts}
\usepackage{amsmath}
\usepackage{amssymb}
\usepackage{mathtools}
\usepackage{amsthm}
\usepackage{bm}
\usepackage{booktabs}
\usepackage{tabularx}
\usepackage{times}
\usepackage{epsfig}
\usepackage{graphicx}
\usepackage{subcaption}
\usepackage{multirow}
\usepackage{makecell}
\usepackage{xcolor}
\usepackage{xspace}
\usepackage{algorithm}
\usepackage{algorithmic}
\usepackage{cuted}
\usepackage[capitalize,noabbrev]{cleveref}
\usepackage[accsupp]{axessibility}

\theoremstyle{plain}

\theoremstyle{definition}

\theoremstyle{remark}

\def\paperID{0000}
\def\confName{CVPR}
\def\confYear{2026}

\newcommand{\ourmethod}{\textsc{SparseGen}\xspace}

\title{Rethinking Image-to-3D Generation with Sparse Queries: \\Efficiency, Capacity, and Input-View Bias}

\makeatletter
\def\@maketitle{
     \newpage
     \null
     \iftoggle{cvprrebuttal}{\vspace*{-.3in}}{\vskip .28in}
     \begin{center}
            \iftoggle{cvprrebuttal}{{\large \bf \@title \par}}{{\Large \bf \@title \par}}
            \iftoggle{cvprrebuttal}{\vspace*{-22pt}}{\vspace*{20pt}}{
                \large
                \lineskip .4em
                \begin{tabular}[t]{c}
                    \iftoggle{cvprfinal}{
                        \@author
                    }{
                        \iftoggle{cvprrebuttal}{}{
                            Anonymous \confName~submission\\
                            \vspace*{1pt}\\
                            Paper ID \paperID
                        }
                    }
                \end{tabular}
                \par
            }
            \vskip .2em
            \vspace*{6pt}
     \end{center}
}
\makeatother

\author{
    Zhiyuan Xu\textsuperscript{\rm 1} \enspace
    Jiuming Liu\textsuperscript{\rm 2} \enspace
    Yuxin Chen\textsuperscript{\rm 1} \enspace
    Masayoshi Tomizuka\textsuperscript{\rm 1} \enspace
    Chenfeng Xu\textsuperscript{\rm 3} \enspace
    Chensheng Peng\textsuperscript{\rm 1} \\[8pt]
    \textsuperscript{\rm 1}UC Berkeley \qquad
    \textsuperscript{\rm 2}University of Cambridge \qquad
    \textsuperscript{\rm 3}UT Austin\\
}

\begin{document}
\maketitle

\begin{strip}
\centering
    \includegraphics[width=0.95\textwidth]{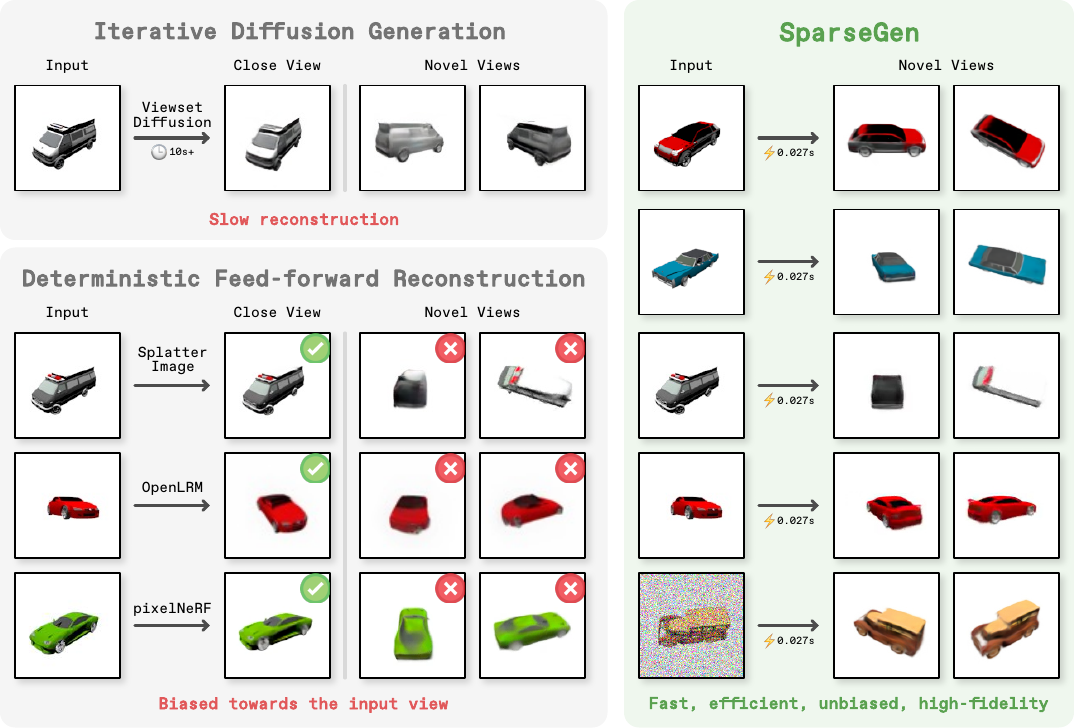}
    \captionof{figure}{\textbf{Comparison with Prior 3D Generation Paradigms.}
    Left: iterative diffusion generation (e.g., Viewset Diffusion) synthesizes multi-view images but requires multiple denoising steps, suffering from poor efficiency.
    Middle: deterministic feed-forward reconstruction methods (e.g., Splatter Image, LRM, pixelNeRF), demonstrate good synthesis quality on views close to the input ones yet degrade on held-out novel viewpoints.
    Right: \ourmethod predicts 3D Gaussians from a sparse set of learned queries, producing high-fidelity, consistent novel views and achieves low input-view bias while being extremely fast.}
    \label{fig:teaser}
\end{strip}

\begin{abstract}

We present \ourmethod, a novel framework for efficient image-to-3D generation, which exhibits low input-view bias while being significantly faster. Unlike traditional approaches that rely on dense volumetric grids, triplanes, or pixel-aligned primitives, we model scenes with a compact sparse set of learned 3D anchor queries and a learned expansion operator that decodes each transformed query into a small local set of 3D Gaussian primitives. Trained under a rectified-flow reconstruction objective without 3D supervision, our model learns to allocate representation capacity where geometry and appearance matter, achieving significant reductions in memory and inference time while preserving multi-view fidelity. We introduce quantitative measures of input-view bias and utilization to show that sparse queries reduce overfitting to conditioning views while being representationally efficient. Our results argue that sparse set-latent expansion is a principled, practical alternative for efficient 3D generative modeling.

\end{abstract}

\section{Introduction}

\label{sec:intro}

Synthesizing photorealistic 3D content from sparse image observations is a fundamental challenge in computer vision and graphics, with applications spanning AR/VR \cite{song2024toward}, robotics simulation \cite{katara2024gen2sim}, and embodied AI \cite{yang2024holodeck}. A key desideratum for such systems is \textbf{low input-view bias}---the ability to produce high-quality novel views from arbitrary viewpoints, not just those well-covered by the conditioning images. This requires modeling the inherent uncertainty and ambiguity in unobserved regions, distinguishing generative 3D synthesis from pure reconstruction tasks.

Meanwhile, recent advances in neural 3D representations, including Neural Radiance Fields (NeRFs)~\cite{mildenhall_NeRFRepresentingScenes_2021} and 3D Gaussian Splatting (3DGS)~\cite{kerbl_3DGaussianSplatting_2023}, have enabled remarkable photorealism. However, many existing approaches face challenges in view consistency and computational efficiency.
\textbf{Deterministic feed-forward methods}~\cite{szymanowicz_SplatterImageUltraFast_2024, hong_LRMLargeReconstruction_2023, yu_PixelNeRFNeuralRadiance_2021, wang_DUSt3RGeometric3D_2024} achieve fast inference by directly mapping input images to 3D representations, 
but often lack generative modeling capacity and degrade significantly on novel viewpoints not well-represented in the input (Figure~\ref{fig:teaser}). Conversely, \textbf{iterative generative methods}~\cite{szymanowicz_ViewsetDiffusion0ImageConditioned_2023, poole_DreamFusionTextto3DUsing_2022, shi_MVDreamMultiviewDiffusion_2023} maintain quality across views through probabilistic modeling, but require dozens to hundreds of denoising steps, resulting in prohibitively slow generation times.
Additionally, most existing methods rely on \emph{dense 3D parameterizations}: voxel grids with millions of cells, point clouds with hundreds of thousands of samples, or dense Gaussian initializations containing tens of thousands of primitives. Such over-parameterized representations incur substantial memory overhead and computational cost, hindering scalability and real-time deployment. This raises a fundamental question: \emph{\textbf{Can we achieve view-unbiased 3D generation with high representation efficiency and fast inference?}}

In this work, we answer this question affirmatively by introducing \ourmethod, a novel 3D generation model that achieves both low input-view bias and exceptional efficiency through \textbf{sparse 3D anchor queries} and a generative framework. It features a 3D position-aware encoder that injects geometric priors, a transformer-based query-to-Gaussian expansion network that decodes sparse anchors into full Gaussian attributes through cross-attention, and differentiable 3DGS rendering enabling end-to-end training with only 2D supervision.
Our method is motivated by the observation that not all 3D locations are equally informative: many voxels represent empty space, numerous points redundantly encode smooth surfaces, and a large fraction of Gaussians in dense initializations contribute negligibly to the final rendering. Therefore, we maintain a small set of learnable 3D anchor queries, where each query token corresponds to a 3D location enriched with learned latent attributes, serving as a \emph{seed} that can be decoded into explicit Gaussians. Crucially, we train these sparse queries within a generative framework, enabling the model to probabilistically infer geometry and appearance in unobserved regions. Unlike prior approaches that rely on dense representations or iterative refinement, \ourmethod permits \textbf{single-step generation}, drastically reducing computational overhead while maintaining generative expressiveness.

Overall, our contributions are:
\begin{itemize}
    
    \item We propose \ourmethod, a sparse query-based 3D generation framework that achieves low input-view bias through generative modeling while being significantly more efficient than iterative diffusion methods.
    
    \item We design a unified architecture combining 3D position-aware encoding, transformer-based query-to-Gaussian expansion, and rectified flow training, enabling single-step feed-forward synthesis from a variable number of input views.
    
    \item We provide comprehensive empirical analysis showing that sparse queries yield high primitive utilization and exhibit query-induced spatial locality, suggesting potential for part-level editing.
    
    \item We demonstrate remarkable quality and efficiency, achieving 600$\times$ speedup compared to iterative baselines while using a compact 280KB representation.
\end{itemize}
\section{Related Works}
\label{sec:related_work}

\subsection{Image-to-3D Generation}

A central challenge in image-to-3D generation is producing high-quality renderings from \emph{arbitrary} novel viewpoints given sparse observations. Optimization-based pipelines that lift 2D priors to 3D, e.g., score distillation sampling (SDS) methods such as DreamFusion~\cite{poole_DreamFusionTextto3DUsing_2022} and related variants~\cite{lin_Magic3DHighResolutionTextto3D_2023, tang_DreamGaussianGenerativeGaussian_2024, liang_LucidDreamerHighFidelityTextto3D_2024}, can generate detailed assets but are computationally expensive and may exhibit geometric inconsistency (e.g., the Janus problem~\cite{shi_MVDreamMultiviewDiffusion_2023}). More recently, generative approaches that explicitly target multi-view consistency via denoising across views~\cite{hollein_ViewDiff3DConsistentImage_2024, zheng_Free3DConsistentNovel_2024} have improved coherence but often inherit the iterative cost of diffusion; for instance, Viewset Diffusion~\cite{szymanowicz_ViewsetDiffusion0ImageConditioned_2023} performs iterative denoising of multi-view images with an inner explicit 3D representation to enable consistent synthesis under 2D supervision, yet its many denoising steps lead to high inference latency.

\subsection{Feed-Forward Reconstruction and Input-View Bias}

A parallel line of work emphasizes fast feed-forward reconstruction. Large Reconstruction Models (LRMs)~\cite{hong_LRMLargeReconstruction_2023, zhang_GSLRMLargeReconstruction_2024, wei_MeshLRMLargeReconstruction_2025a}, Splatter Image~\cite{szymanowicz_SplatterImageUltraFast_2024}, 3Rs~\cite{wang_Continuous3DPerception_2025,wang_DUSt3RGeometric3D_2024, leroy_GroundingImageMatching_2025, yang_Fast3R3DReconstruction_2025} map one or a few images to a 3D representation in a single forward pass. While being more efficient, purely deterministic mappings are often \emph{input-view biased}: they tend to perform best on viewpoints close to the conditioning views and may degrade on held-out views due to the lack of an explicit modeling for ambiguous, unobserved regions, as do some works that unify 3d reconstruction and rendering with a large transformer~\cite{jin_LVSMLargeView_2024, sajjadi_SceneRepresentationTransformer_2022}.

\subsection{Efficient Representations, Sparse Queries, and 3D Gaussian Splatting}

3D representations trade off fidelity, efficiency, and scalability. Point clouds~\cite{hoppe_SurfaceReconstructionUnorganized_1992} and meshes~\cite{catmull_ComputerDisplayCurved_1998} are compact but can be challenging to render or generate robustly from sparse inputs, while voxel grids~\cite{brebin_VolumeRendering_1998} are straightforward but scale cubically in memory and compute.
Neural Radiance Fields (NeRFs)~\cite{mildenhall_NeRFRepresentingScenes_2021} achieve high photorealism but require expensive per-ray sampling for rendering. In contrast, 3D Gaussian Splatting (3DGS)~\cite{kerbl_3DGaussianSplatting_2023} represents scenes with explicit Gaussian primitives and enables fast differentiable rasterization, making it well-suited for efficient learning and real-time novel-view synthesis.

Orthogonal to the representation choice, sparse \emph{query} or \emph{set-latent} modeling provides a principled capacity bottleneck: learned queries summarize inputs and are decoded to structured outputs, as popularized by DETR~\cite{carion_EndtoEndObjectDetection_2020} and extended to 3D reasoning with 3D queries in multi-view settings~\cite{wang_DETR3D3DObject_2022, liu_PETRPositionEmbedding_2022}. Inspired by this paradigm, we model a scene with a small set of learned 3D anchor queries and decode them into compact 3DGS primitives, enabling efficient capacity allocation and fast inference while maintaining view-consistent generation.

\begin{figure*}[t]
    \centering
    \includegraphics[width=\linewidth]{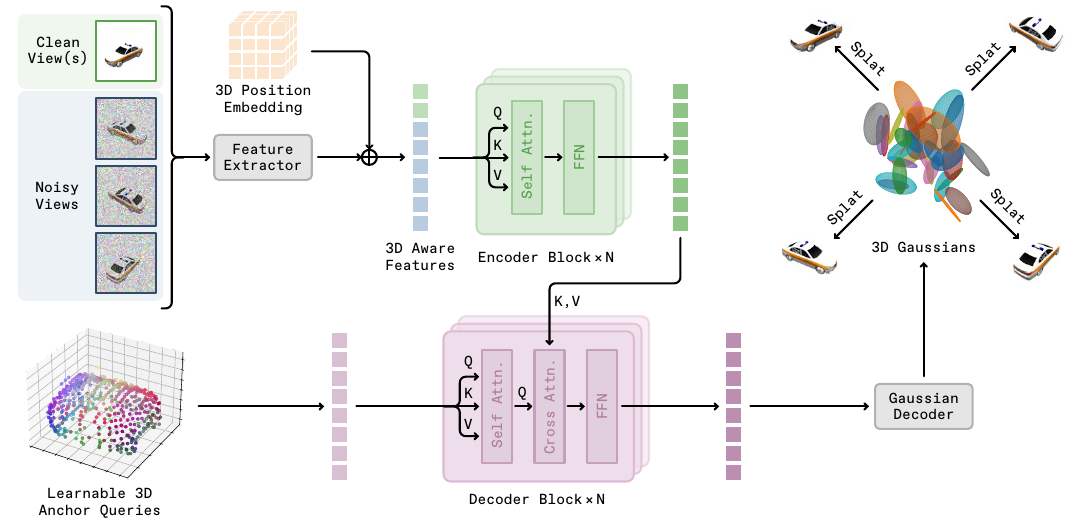}
    \caption{\textbf{Overview of \ourmethod.} Given $V$ input views (clean and/or noisy) with known camera poses, an image encoder (with adaLN timesteps) and a 3D position encoder generate position-aware image features. A sparse set of learnable 3D anchor queries attends to these fused features in a transformer-based expansion network and is decoded into a compact set of 3D Gaussians. Finally, the generated Gaussians are rendered for target views via differentiable splatting, enabling fast, high-quality 3D generation and rendering.}
    
    \label{fig:method_overview}
\end{figure*}

\section{Method}

Following Viewset Diffusion~\cite{szymanowicz_ViewsetDiffusion0ImageConditioned_2023}, we formulate the 3D generation task by synthesizing a set of multi-view images that are rendered from 3D Gaussian representations and supervised by ground truth images. However, unlike Viewset Diffusion which relies on an iterative denoising process to gradually refine multi-view images and the 3D representation during inference, our method employs an efficient query expansion network which is trained under the rectified flow paradigm~\cite{lipman_FlowMatchingGenerative_2022a, liu_FlowStraightFast_2022}, thereby facilitating high-quality 3D Gaussian generation in a one-step feed-forward pass. An overview of our method is illustrated in \Cref{fig:method_overview}.

\subsection{Preliminaries}

\noindent\textbf{Rectified Flow.} Rectified Flow~\cite{lipman_FlowMatchingGenerative_2022a, liu_FlowStraightFast_2022} is a generative modeling framework that maps Gaussian noise to data samples via straight paths in the data space as:
\begin{equation}
    \label{eq:rectified_flow}
    x_t = (1 - t) x_0 + t \epsilon, \quad t \in [0, 1]
\end{equation}
where $x_0$ is a data sample, $\epsilon \sim \mathcal{N}(0, I)$ is Gaussian noise, and $x_t$ is the interpolated noisy sample at time $t$.
Typically, a neural network $v_\theta(x_t, t)$ is trained to predict the velocity from $x_t$ to $x_0$, but our implementation predicts the denoised sample $x_0$ instead since the inner Gaussian representation directly models the clean sample without noise.

\noindent\textbf{3D Gaussian Splatting.} 3D Gaussian Splatting~\cite{kerbl_3DGaussianSplatting_2023} represents a scene using a set of colored, anisotropic Gaussian primitives $g_i = (\mu_i, \Sigma_i, c_i, \alpha_i)$, where $\mu_i \in \mathbb{R}^3$ is the 3D mean, $\Sigma_i$ is the covariance matrix controlling the shape and orientation, $c_i$ is the color, and $\alpha_i$ is the opacity. Each primitive is projected to the image plane as an ellipse with mean $x_i = \pi(\mu_i)$ and covariance $\Sigma_i^{\text{img}}$. For each pixel $p$, the rendered color is computed by front-to-back alpha compositing:
\begin{align}
C(p) &= \sum_i \Big(\prod_{j<i} (1 - w_j(p))\Big) w_i(p) c_i,\\
w_i(p) &= \alpha_i \exp\big(-\tfrac{1}{2}(p - x_i)^\top (\Sigma_i^{\text{img}})^{-1} (p - x_i)\big).
\end{align}
where $w_i(p)$ denotes the pixel-wise contribution of the $i$-th Gaussian to pixel $p$.
This closed-form differentiable rasterization eliminates the need for per-ray sampling, providing high rendering efficiency and stable gradient propagation, making it ideal for our feed-forward generative formulation.

\begin{figure}[t]
    \centering
    \includegraphics[width=\linewidth]{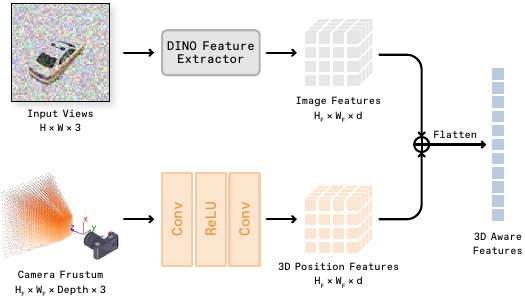}
    \caption{\textbf{3D Position-Aware Encoder.} Pixels are unprojected along the camera frustum to fixed depths and mapped by a small Conv--ReLU--Conv into per-pixel 3D features. These 3D features are merged with DINO-extracted image features and flattened into 3D position-aware tokens for the following transformer.}
    \label{fig:3dpos_encoder}
    
\end{figure}

\subsection{The \textbf{\ourmethod} Model}

\Cref{fig:method_overview} illustrates the overall architecture of our model. Given a set of input images $x_{t_1, t_2, \ldots, t_V} \in \mathbb{R}^{V \times H \times W \times 3}$ (either clean or noisy, with known camera poses $\pi_{1, 2, \ldots, V}$), the model generates a set of 3D Gaussians $\mathcal{G} = \{g_i\}_{i=1}^N$ representing the underlying 3D scene and renders them into corresponding clean images $\hat{x}_{0, 0, \ldots, 0}$.

\noindent\textbf{Image Feature Extraction.} We adapt a DINOv2-like~\cite{oquab_DINOv2LearningRobust_2024} architecture to extract image feature tokens, with added adaptive layer normalization~\cite{peebles_ScalableDiffusionModels_2023} to accept timestep $t$ as input, which indicates the noise level of the images. This module transforms input images $x_{t_1, t_2, \ldots, t_V}\in \mathbb{R}^{V \times H \times W \times 3}$ into feature tokens $F_I \in \mathbb{R}^{V \times H_F \times W_F \times d}$ where $d$ is the feature dimension.

\noindent\textbf{3D Positional Embedding.} To inject 3D spatial information into the extracted 2D image features, we employ a 3D position encoder as illustrated in \Cref{fig:3dpos_encoder}. We first unproject each image pixel into 3D space based on its camera parameters at fixed depth intervals, obtaining a frustum of 3D points $P \in \mathbb{R}^{V \times H_F \times W_F \times dth \times 3}$ where $dth$ is the number of depth samples. These 3D points are then encoded using a 1×1 convolutional neural network to align with the feature dimension $d$ from image features, resulting in 3D positional embeddings $F_P \in \mathbb{R}^{V \times H_F \times W_F \times d}$. $F_P$ and $F_I$ are then merged to produce 3D position-aware features $F_{3D}$.

\noindent\textbf{Query-to-Gaussian Expansion Network.} We maintain a set of learnable 3D anchor queries $Q \in \mathbb{R}^{M \times d}$ where $M$ is the number of queries, each representing a coarse anchor in 3D space. These queries are expanded into a full set of 3D Gaussians via a transformer-based expansion network. In this network, 3D position-aware image features $F_{3D}$ are first flattened and passed through several transformer encoder layers with self-attention to aggregate multi-view context. The anchor queries $Q$ then attend to these features through decoder layers with cross-attention, allowing them to gather relevant information from the images. 
Finally, an MLP-based Gaussian head decodes the output query features into Gaussian parameters: mean $\mu_i$, covariance $\Sigma_i$, color $c_i$, and opacity $\alpha_i$ for each Gaussian $g_i$, where each query generates a fixed number of Gaussians.

Formally, we summarize the unified forward routine and its use during training and inference in Algorithm~\ref{alg:method}; see \Cref{sec:training}, \Cref{sec:inference} and \Cref{sec:impl_details} for details. The same forward pass is shared in both settings. Training differs only in view/noise sampling and loss computation.

\begin{algorithm}[t]
\caption{Training and inference for \ourmethod}
\label{alg:method}
\footnotesize
\begin{algorithmic}[1]
\STATE \textbf{procedure} \textsc{Forward}($x_{t[1:V]}, t_{[1:V]}, \pi_{[1:V]}$)
\STATE \quad $F_I \gets \text{ImageEncoder}(x_{t[1:V]}, t_{[1:V]})$
\STATE \quad $F_P \gets \text{3DPosEncoder}(\pi_{[1:V]})$ \COMMENT{frustum unprojection}
\STATE \quad $F_{3D} \gets \text{Flatten}(F_I + F_P)$
\STATE \quad \textbf{for} $i=1 \ldots N_{\text{enc}}$ \textbf{do}
\STATE \quad\quad $F_{3D} \gets \text{SelfAttn}(Q{=}F_{3D}, K{=}F_{3D}, V{=}F_{3D})$

\STATE \quad \textbf{end for}
\STATE \quad $Q \gets \text{LearnableParameters}$ \COMMENT{learnable 3D anchor queries}
\STATE \quad \textbf{for} $i=1 \ldots N_{\text{dec}}$ \textbf{do}
\STATE \quad\quad $Q \gets \text{CrossAttn}(Q{=}\text{SelfAttn}(Q), K{=}F_{3D}, V{=}F_{3D})$

\STATE \quad \textbf{end for}
\STATE \quad $\mathcal{G} \gets \text{GaussianHead}(Q)$ \COMMENT{$\{\mu, \Sigma, c, \alpha\}$}
\STATE \quad $\hat{x}_0 \gets \text{Render3DGS}(\mathcal{G}, \pi_{[1:V]})$ \COMMENT{differentiable splatting}
\STATE \quad \textbf{return} $(\mathcal{G}, \hat{x}_0)$
\STATE \textbf{end procedure}
\STATE
\STATE \textbf{procedure} \textsc{TrainingStep}(batch)
    \STATE \quad Select $V$ views with poses $\pi_{[1:V]}$
    \STATE \quad Sample $t \sim \mathcal{U}[0,1]$; add noise $x_t = (1-t)x_0 + t\,\epsilon$
    \STATE \quad Optionally drop a subset of views (robustness)
    \STATE \quad $(\mathcal{G}, \hat{x}_0) \gets \textsc{Forward}(x_{t[1:V]}, t_{[1:V]}, \pi_{[1:V]})$
    \STATE \quad Compute $L = L_2(\hat{x}_0, x_0) + L_{\text{perc}} + L_{\text{opacity}} + L_{\text{gauss\_reg}}$
    \STATE \quad Backpropagate and update parameters
    \STATE \quad \textbf{return} $L$
\STATE \textbf{end procedure}
\STATE
\STATE \textbf{procedure} \textsc{Inference}(conditioning\_views)
    \STATE \quad Concat clean conditioning views with noise placeholders
    \STATE \quad Set $t=0$ for clean views, $t=1$ for noise placeholders
    \STATE \quad $(\mathcal{G}, \hat{x}_0) \gets \textsc{Forward}(x_{t[1:V]}, t_{[1:V]}, \pi_{[1:V]})$
    \STATE \quad \textbf{return} $(\mathcal{G}, \hat{x}_0)$ \COMMENT{Gaussians \& rendered novel views}
\STATE \textbf{end procedure}
\end{algorithmic}
\end{algorithm}

\subsection{Training}

\label{sec:training}
\noindent\textbf{Training Data.} To train \ourmethod, we need a dataset consisting of multi-view RGB images of 3D objects with known camera poses, with optional alpha~\cite{porter_CompositingDigitalImages_1984} masks for foreground-background separation. Explicit 3D information, such as point maps, is not required. For each training sample, we randomly select $V$ views from all available images of the object as input, and add Gaussian noise to some of them based on a random timestep $t$ and the rectified flow formulation, as illustrated in \Cref{eq:rectified_flow}.

\noindent\textbf{Loss Function.} We train the model end-to-end with the image reconstruction loss between rendered images $\hat{x}_{0, 0, \ldots, 0}$ (denoted as $\hat{x}_0$ for simplicity) and the ground-truth clean images $x_0$. Specifically, we use a combination of L2 loss and perceptual loss~\cite{johnson_PerceptualLossesRealTime_2016} to encourage both pixel-level accuracy and perceptual quality:
\begin{equation}
    \begin{split}
        \mathcal{L}_{rec} = \lambda_{L2} \| \hat{x}_0 - x_0 \|_2^2 + 
            \lambda_{perc} \mathcal{L}_{perc}(\hat{x}_0, x_0),
    \end{split}
\end{equation}
where $\lambda_{L2}$ and $\lambda_{perc}$ are weighting factors, together with an optional L2 loss on opacity values (when alpha masks are available).
Moreover, we add regularization terms to the Gaussian parameters to promote reasonable distributions. For more information, we refer the reader to \Cref{sec:impl_details}.

\noindent\textbf{Training Procedure.} During training, we randomly sample 5 views and add Gaussian noise of equivalent strength to 3 of them based on a random timestep $t$. Before feeding the images into the model, we randomly drop out some input views (could be both noisy and clean ones), but still supervise the model to reconstruct all 5 clean views. This encourages the model to be robust to varying numbers of input views and noise levels, and to effectively leverage multi-view context for accurate 3D Gaussian generation.

\subsection{Inference}

\label{sec:inference}
During inference, \ourmethod can generate 3D Gaussians in a one-step feed-forward pass. Given a clean conditioning image, we concatenate it with randomly sampled Gaussian noise images and feed them into the model to generate 3D Gaussians. Note that our model naturally supports varying numbers of input views, or pure noisy inputs for unconditional generation, without any architecture changes. The generated 3D Gaussians can be directly rendered into novel views using the 3DGS differentiable renderer for high-fidelity, consistent multi-view synthesis.\looseness=-1
\section{Experiments}

\begin{table*}[t]
\centering
\caption{\textbf{Single-view reconstruction on ShapeNet-SRN Cars.} One conditioning view is provided; metrics are averaged over the remaining ground-truth views. Higher is better for PSNR/SSIM; lower is better for LPIPS/FID. ``Reconstruction Time'' is the wall-clock time to produce the 3D representation on a single NVIDIA L40; ``3D Representation Size'' reports memory footprint and the number/type of primitives. Best results are in \textbf{bold} while runners-up are \underline{underlined}.
\ourmethod uses 5{,}120 Gaussians and achieves the best PSNR and FID with the fastest runtime and smallest footprint. Note that Splatter Image uses degree 1 spherical harmonics for color representation, while \ourmethod uses RGB colors, and therefore the number of parameters per Gaussian is different.}

\resizebox{\textwidth}{!}{
\begin{tabular}{l|cccc|cl}
    \toprule
    Method            & PSNR $\uparrow$  & LPIPS $\downarrow$ & SSIM $\uparrow$  & FID $\downarrow$    & Reconstruction Time $\downarrow$  & 3D Representation Size $\downarrow$  \\
    \midrule
    OpenLRM~\cite{hong_LRMLargeReconstruction_2023}           & 18.286 & 0.134 & 0.815 & 51.421 & 0.301s           & 3,840KB (12,288 Triplane Cells)         \\
    Splatter Image~\cite{szymanowicz_SplatterImageUltraFast_2024}    & \underline{23.933} & \textbf{0.077} & \textbf{0.922} & 44.908 & \underline{0.042s}           & \underline{1,472KB} (16,384 Gaussians)      \\
    Viewset Diffusion~\cite{szymanowicz_ViewsetDiffusion0ImageConditioned_2023} & 22.688 & 0.096 & 0.891 & \underline{39.807} & 16.32s           & 8,192KB (32,768 Voxel Cells)   \\
    \midrule
    \ourmethod        & \textbf{24.018} & \underline{0.081} & \underline{0.913} & \textbf{23.595} & \textbf{0.027s}           & \textbf{280KB} (5,120 Gaussians)       \\
    \bottomrule
\end{tabular}
}

\label{tab:shapenet_srn_results}

\end{table*}

\subsection{Experimental Setup}

\noindent\textbf{Datasets.} We adopt ShapeNet-SRN~\cite{sitzmann_SceneRepresentationNetworks_2019} as our primary dataset for 3D object generation, with standard train/val/test splits.
For each test object, one clean view is provided as conditioning input, and the model is expected to generate the remaining 250 novel views, which are compared against ground-truth images for quantitative evaluation. We also conduct experiments on the CO3D~\cite{reizenstein_CommonObjects3D_2021} dataset, with a test split of 100 objects each category, where one view serves as input and another view is held out for evaluation.
Additionally, larger datasets are used to test the potential of \ourmethod in \Cref{sec:add_res}.

\noindent\textbf{Evaluation Metrics.} We evaluate the quality of generated multi-view images using standard image metrics including Peak Signal-to-Noise Ratio (PSNR)~\cite{10.5555/549677}, Structural Similarity Index Measure (SSIM)~\cite{wang_ImageQualityAssessment_2004}, and Learned Perceptual Image Patch Similarity (LPIPS)~\cite{zhang_UnreasonableEffectivenessDeep_2018}. Additionally, we compute Fréchet Inception Distance (FID)~\cite{heusel_GANsTrainedTwo_2017a} to assess the overall realism of generated images. For the ShapeNet-SRN dataset, we randomly sample 15 views per object from the generated and ground-truth sets, totaling around 10k images per set for reliable FID computation. For the CO3D dataset, we do not compute FID due to the limited number of test images. We also measure inference speed to demonstrate the efficiency of our approach, reporting the average time taken to generate the 3d representations with a single NVIDIA L40 GPU.

\begin{table}[t]
\centering
\caption{\textbf{Single-view reconstruction on CO3D subsets.} One input view per object; metrics computed on a held-out target view. Results are reported per category. \ourmethod improves PSNR/LPIPS/SSIM over Viewset Diffusion on both Hydrant and Teddybear.}

\resizebox{\columnwidth}{!}{
\begin{tabular}{l|l|ccc}
    \toprule
Subset     & Method            & PSNR $\uparrow$   & LPIPS $\downarrow$ & SSIM $\uparrow$ \\
\midrule
\multirow{2}{*}{Hydrant}  & Viewset Diffusion & 19.664 & 0.232 & 0.693 \\
                                & \ourmethod        & \textbf{20.366} & \textbf{0.192} & \textbf{0.724} \\
\midrule
\multirow{2}{*}{Teddybear} & Viewset Diffusion & 15.473 & 0.405 & 0.492 \\
                                & \ourmethod        & \textbf{19.005} & \textbf{0.353} & \textbf{0.568} \\
    \bottomrule
\end{tabular}

}

\label{tab:co3d_subsets}

\end{table}

\noindent\textbf{Compared Baselines.} We primarily compare \ourmethod against Viewset Diffusion~\cite{szymanowicz_ViewsetDiffusion0ImageConditioned_2023}, as our method shares the same diffusion-based generative paradigm but with a sparse query representation. We also include comparisons with recent feed-forward reconstruction methods including Splatter Image~\cite{szymanowicz_SplatterImageUltraFast_2024} and OpenLRM~\cite{hong_LRMLargeReconstruction_2023}, to highlight the advantages of our generative approach over these deterministic reconstruction methods. More recent methods that require substantially larger compute to train are not included.

\subsection{Reconstruction Results}

\noindent\textbf{Single-view Reconstruction on the ShapeNet-SRN dataset.}
We evaluate the single-view reconstruction performance on ShapeNet-SRN Cars in \Cref{tab:shapenet_srn_results}.

Owing to our sparse query-based model design, it only needs 0.027s to reconstruct an object, which is over 600$\times$ faster than the iterative Viewset Diffusion \cite{szymanowicz_ViewsetDiffusion0ImageConditioned_2023}. Despite the small number of Gaussians used (5{,}120 Gaussians, decoded from 512 output tokens) and a compact 280KB representation, \ourmethod attains the best PSNR and the lowest FID, indicating higher reconstruction quality and more realistic novel views. LPIPS and SSIM of \ourmethod are also competitive, achieving second-best and on par with the feed-forward baselines respectively. The superiority of \ourmethod in quality, speed, and compactness highlights the advantage of our sparse Gaussian representations.

\noindent\textbf{Two-view Reconstruction on the ShapeNet-SRN dataset.}

\Cref{tab:two_view_srn} reports the two-view reconstruction performance on ShapeNet-SRN Cars,
which shows that \ourmethod obtains competitive results across all metrics with the best efficiency. Note that the results we reproduced for Splatter Image~\cite{szymanowicz_SplatterImageUltraFast_2024} with their released checkpoints are lower than those reported in their paper, likely because they trained separate models for single- and two-view settings, whereas our method does not require per-view-count retraining. Furthermore, our method maintains a constant 3D representation size regardless of the number of input views, while Splatter Image's representation size grows linearly with the number of input views since it predicts Gaussians per input pixel.

\noindent\textbf{Single-view Reconstruction on the CO3D dataset.} We further conduct experiments on the CO3D dataset to evaluate 3D generation performance. Results are summarized in \Cref{tab:co3d_subsets}. Compared to Viewset Diffusion, the primary generative baseline, \ourmethod achieves significant improvements across all three metrics on both Hydrant and Teddybear categories, demonstrating better generalization.

\begin{figure}[t]
    \centering
    \includegraphics[width=\linewidth]{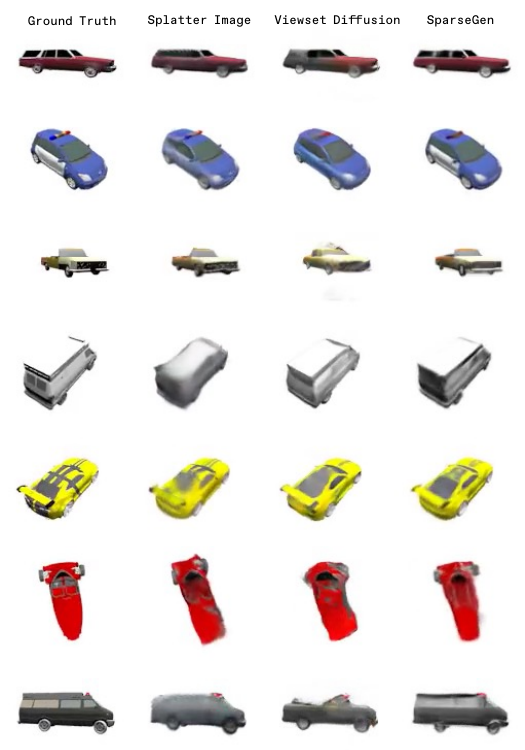}
    
    \caption{\textbf{Qualitative results on ShapeNet-SRN Cars.} We show examples under one-view conditioning. \ourmethod yields sharper details, cleaner boundaries, with fastest generation speed.}
    \label{fig:quali_srn}
    
\end{figure}

\noindent\textbf{Qualitative Results.} \Cref{fig:quali_srn} shows qualitative comparisons on ShapeNet-SRN Cars under single-view conditioning. Additional results on CO3D are provided in the appendix (\Cref{fig:quali_hyd,fig:quali_ted,fig:obj_quali}).

\begin{table}[t]
\centering
\caption{\textbf{Ablations on ShapeNet-SRN Cars.} Each row removes one component from \ourmethod. Rectified flow improves realism and fidelity; 3D positional embeddings inject essential geometry; learnable 3D anchor queries are critical—removing them causes a quality drop and a large FID increase.}

\resizebox{\columnwidth}{!}{
\begin{tabular}{l|cccc}
    \toprule
Method                          & PSNR $\uparrow$   & LPIPS $\downarrow$ & SSIM $\uparrow$  & FID $\downarrow$    \\
    \midrule

w/o rectified flow              & 23.069 & 0.092 & 0.895 & 28.415  \\
w/o 3d pos embed    & 20.757 & 0.108 & 0.861 & 27.946  \\
w/o learnable queries & 17.159 & 0.209 & 0.807 & 178.129 \\
\midrule
\ourmethod                      & \textbf{24.018} & \textbf{0.081} & \textbf{0.913} & \textbf{23.595}  \\
    \bottomrule
\end{tabular}
}
\label{tab:ablation_studies}

\end{table}

\begin{table*}[t]
\centering
\caption{\textbf{Two-view reconstruction on ShapeNet-SRN Cars.} Higher is better for PSNR/SSIM; lower is better for LPIPS/FID. ``Representation Size'' indicates the relative size of the 3D representation compared to, when using one view as input. We evaluated baselines with their released checkpoints. Best results are in \textbf{bold}.}

\resizebox{0.9\linewidth}{!}{
\begin{tabular}{l|cccc|cc}
    \toprule
Method            & PSNR $\uparrow$  & LPIPS $\downarrow$ & SSIM $\uparrow$  & FID $\downarrow$ & Reconstruction Time $\downarrow$  & Representation Size $\downarrow$ \\
\midrule
Viewset Diffusion ~\cite{szymanowicz_ViewsetDiffusion0ImageConditioned_2023} & 25.053 & 0.079 & 0.919 & 35.934 & 21.26s & 1$\times$                  \\

Splatter Image ~\cite{szymanowicz_SplatterImageUltraFast_2024}   & 23.257 & 0.075 & 0.916 & 40.957 & 0.052s & 2$\times$                  \\
\midrule
\ourmethod        & \bf25.982 & \bf0.065 & \bf0.935 & \bf19.536 & \bf0.029s & 1$\times$                 \\
    \bottomrule
\end{tabular}
}

\label{tab:two_view_srn}

\end{table*}

\subsection{Ablation Studies}

We investigate the effectiveness and contribution of each component in \Cref{tab:ablation_studies}. First, replacing rectified flow with a deterministic direct mapping slightly degrades all metrics, underscoring the value of a generative path in handling single-view ambiguity. Next, removing 3D positional embeddings causes a larger drop in reconstruction fidelity, showing that injecting explicit 3D spatial context into 2D features is important for consistent cross-view reasoning. Finally, substituting the learnable 3D anchor queries with fixed random ones produces the largest decline across PSNR, SSIM, LPIPS, and FID, revealing that these anchors furnish critical spatial priors that guide coherent Gaussian generation. In summary, sparse learnable anchors provide the core generative scaffold, while rectified flow and 3D positional encoding jointly stabilize and refine quality.

\begin{table}[t]
    \centering
    \caption{\textbf{Input-view bias analysis.} Gap between conditioning and novel views on ShapeNet-SRN Cars. \emph{Larger absolute values} indicate \emph{stronger bias.}}
    
    \resizebox{0.8\columnwidth}{!}{
        \begin{tabular}{l|rrr}
            \toprule
            Method            & $\Delta$PSNR      & $\Delta$LPIPS      & $\Delta$SSIM      \\
            \midrule
            OpenLRM           & 8.819             & -0.103             & 0.138             \\
            Splatter Image    & 14.821            & -0.072             & \underline{0.073} \\
            Viewset Diffusion & \underline{8.445} & \underline{-0.061} & 0.084             \\
            \ourmethod        & \textbf{3.502}    & \textbf{-0.025}    & \textbf{0.041}    \\
            \bottomrule
        \end{tabular}
    }
    \label{tab:input_view_bias}
    
\end{table}

\begin{figure}[t]
    \centering
    \includegraphics[width=0.95\columnwidth]{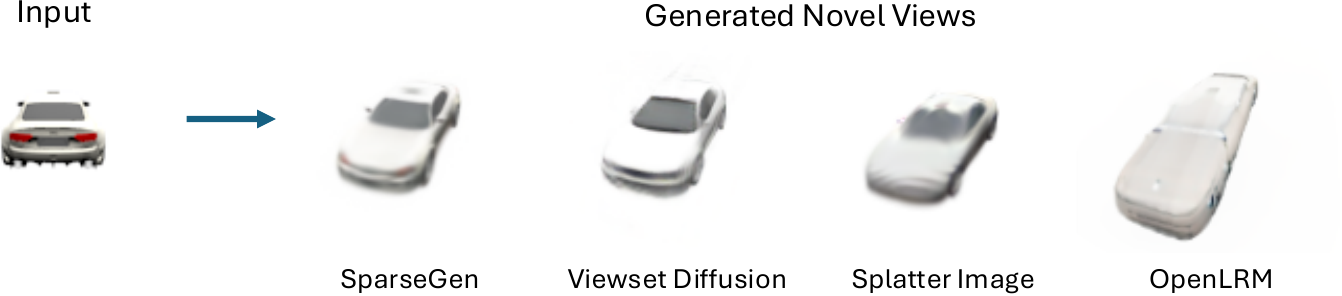}
    \caption{\textbf{Back-view conditioning qualitative example.} The input view observes the object from the back, providing limited information about the front. \ourmethod produces plausible novel views, while deterministic feed-forward baselines fail.}
    \label{fig:back_gen}
    
\end{figure}

\subsection{Input-View Bias}

While image-to-3D methods are often evaluated by aggregating metrics over all test views, this can hide an important phenomenon: many methods perform much worse on novel viewpoints compared to conditioning views.
Following the discussion in \Cref{sec:related_work}, we refer to this phenomenon as \emph{input-view bias}. Intuitively, deterministic regressors can overfit to the visible surface regions and appearance statistics in the conditioning image(s), while struggling to hallucinate occluded geometry and textures for unseen regions.

To quantify this bias, we calculate metrics separately on conditioning views and held-out novel views, then compute the gaps between them as $\Delta m = m_{\text{cond}} - m_{\text{novel}}$, where $m_{\text{cond}}$ and $m_{\text{novel}}$ denote the metric averaged over the conditioning and novel view sets, respectively.\looseness=-1

Results are presented in \Cref{tab:input_view_bias}. As expected, deterministic feed-forward methods (OpenLRM~\cite{hong_LRMLargeReconstruction_2023} and Splatter Image~\cite{szymanowicz_SplatterImageUltraFast_2024}) exhibit significantly larger gaps, while \ourmethod achieves the smallest gaps across all metrics, which indicates more view-unbiased generation: it maintains comparable fidelity on viewpoints far from the conditioning view(s) while preserving strong performance near the input.

This effect is also evident qualitatively: under back-view conditioning where the front side is largely unobserved, deterministic baselines often fail to generate a reasonable front view, whereas \ourmethod can synthesize plausible novel views (\Cref{fig:back_gen}).

\begin{figure}[t]
    \centering
    \begin{subfigure}[t]{0.49\linewidth}
        \centering
        \includegraphics[width=\linewidth]{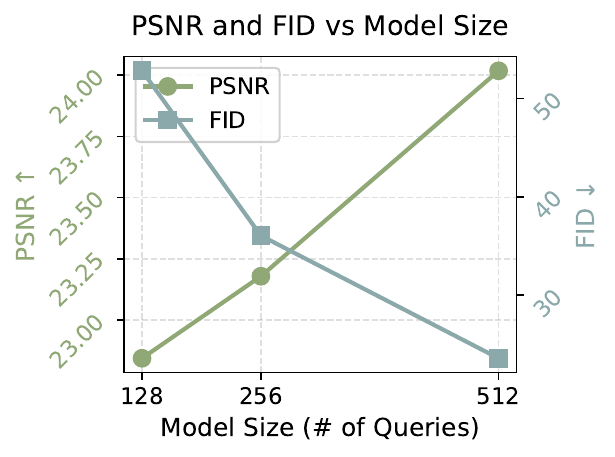}
        
        \caption{PSNR/FID vs. $M$}
    \end{subfigure}
    \hfill
    \begin{subfigure}[t]{0.49\linewidth}
        \centering
        \includegraphics[width=\linewidth]{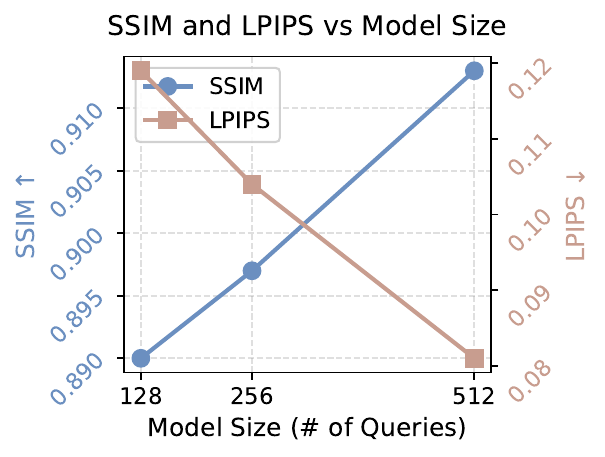}
        
        \caption{SSIM/LPIPS vs. $M$}
    \end{subfigure}
    
    \caption{\textbf{Scalability with respect to the number of queries.} Increasing the number of learned 3D anchor queries ($M$) consistently improves reconstruction quality, supporting the view that queries provide a principled, low-waste representation bottleneck.}
    \label{fig:query_scaling}
    
\end{figure}

\begin{figure}[t]
    \centering
    \includegraphics[width=\linewidth]{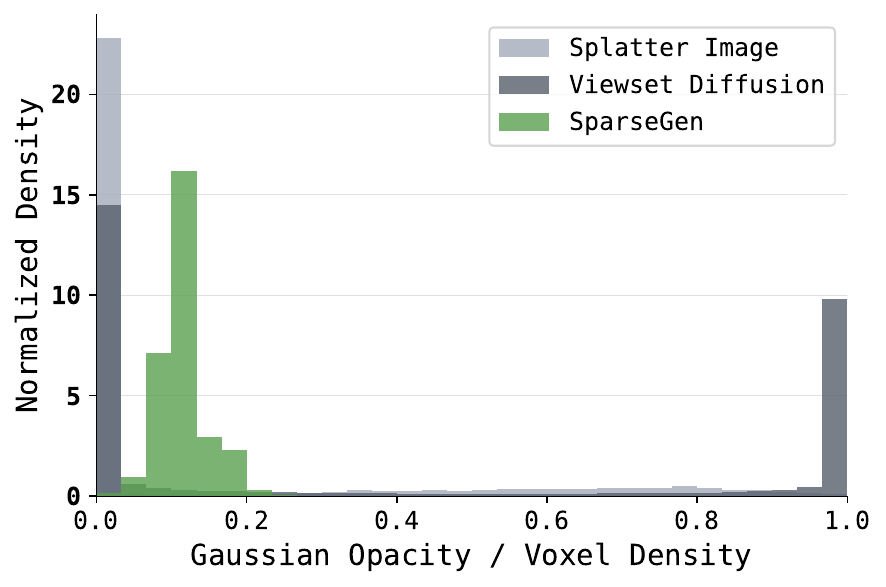}
    
    \caption{\textbf{Opacity/density utilization.} Histogram comparison across methods. Splatter Image produces many near-transparent Gaussians, whereas \ourmethod yields predominantly non-trivial opacities. Viewset Diffusion has many zero-density voxels. For Viewset Diffusion, densities are passed through a sigmoid and rescaled to [0,1] for visualization.}
    \label{fig:density_comparison}
    
\end{figure}

\subsection{Representation Scaling and Utilization}

\label{sec:rep_util}

From a representation perspective, \ourmethod models each object with a compact set of $M$ learned 3D anchor queries (tokens), where each query expands into a small fixed number of 3D Gaussian primitives. This makes $M$ a principled capacity and compute knob: increasing $M$ allocates more representational budget, while keeping the representation explicit and sparse.

\noindent\textbf{Scaling with number of queries.} We run a minimal scaling experiment that varies the number of anchor queries $M$ while keeping the overall architecture unchanged. \Cref{fig:query_scaling} shows that reconstruction quality improves smoothly as $M$ increases (higher PSNR/SSIM and lower LPIPS/FID), indicating that the query set indeed functions as the primary representation bottleneck and that additional queries translate into effective capacity rather than redundant tokens.

\begin{figure}[t]
    \centering
    \begin{subfigure}[t]{0.3\linewidth}
        \centering
        \includegraphics[width=\linewidth]{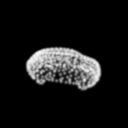}
        \caption{Mean projection per query}
        \label{fig:query_vis_a}
    \end{subfigure}
    \hfill
    \begin{subfigure}[t]{0.3\linewidth}
        \centering
        \includegraphics[width=\linewidth]{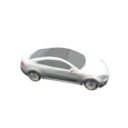}
        \caption{Corresponding RGB image}
        \label{fig:query_vis_b}
    \end{subfigure}
    \hfill
    \begin{subfigure}[t]{0.3\linewidth}
        \centering
        \includegraphics[width=\linewidth]{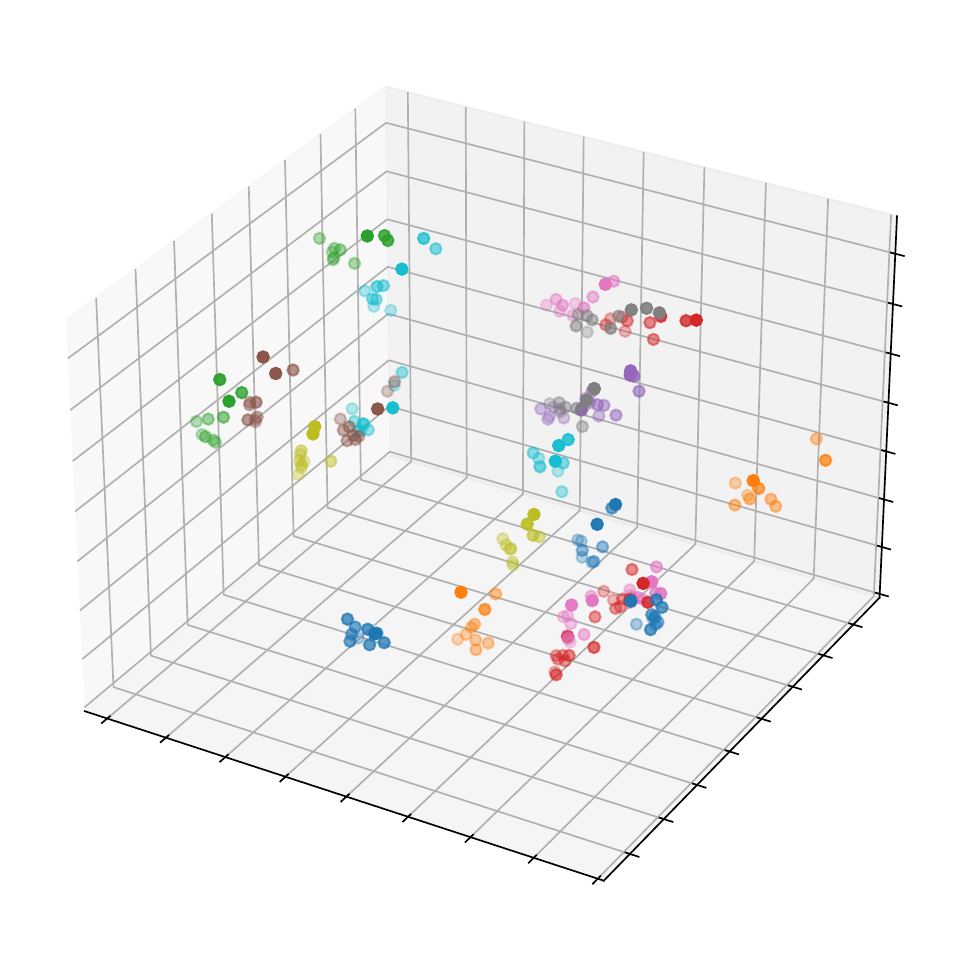}
        \caption{Query-induced locality}
        \label{fig:query_vis_c}
    \end{subfigure}
    
    \caption{\textbf{Qualitative visualization of query utilization and locality.} (a) Projection of per-query mean decoded Gaussian centers onto the image plane. (b) Corresponding RGB image. (c) A subset of decoded Gaussian centers, colored by the anchor query that generated them, illustrating query-induced locality.}
    \label{fig:query_vis}
    
\end{figure}

\noindent\textbf{Utilization of representation.} Beyond scaling, we study whether the allocated primitives are actually used. \Cref{fig:density_comparison} compares the opacity/density distributions across methods. Splatter Image~\cite{szymanowicz_SplatterImageUltraFast_2024} exhibits a long tail of near-transparent Gaussians, suggesting many primitives contribute negligibly and thus waste compute/memory, while Viewset Diffusion~\cite{szymanowicz_ViewsetDiffusion0ImageConditioned_2023} shows a large mass of empty (zero-density) voxels. In contrast, \ourmethod expands most queries into Gaussians with non-trivial opacity, indicating high utilization and explaining the strong quality--size--speed trade-off in \Cref{tab:shapenet_srn_results}. We further visualize utilization by projecting, for each anchor query, the average decoded Gaussian center onto the image plane (\Cref{fig:query_vis_a}) alongside the RGB image (\Cref{fig:query_vis_b}); the projected centers concentrate on object regions, suggesting minimal representational waste.

\noindent\textbf{Query-induced locality and potential editability.} Finally, we visualize the spatial structure induced by the query-to-Gaussian expansion. \Cref{fig:query_vis_c} shows that Gaussians decoded from the \emph{same} anchor query tend to be spatially close, suggesting that each query often captures a coherent local part. This aligns with our motivation of using sparse 3D anchors to structure generation, and suggests a promising direction for future work: query-level 3D editing (e.g., manipulating a subset of queries to edit a semantic part).

\subsection{Training Efficiency}

We report rough training computational costs with GPU-days to contextualize cost across methods. \ourmethod is trained with a budget comparable to Viewset Diffusion~\cite{szymanowicz_ViewsetDiffusion0ImageConditioned_2023} and Splatter Image~\cite{szymanowicz_SplatterImageUltraFast_2024}, while being orders of magnitude less expensive than large-capacity models such as LRM~\cite{hong_LRMLargeReconstruction_2023} and LVSM~\cite{jin_LVSMLargeView_2024}.
Although absolute numbers vary with hardware and schedules, the relative trend is consistent, demonstrating the training efficiency of our proposed method.

\begin{table}[t]
    \centering
    \caption{\textbf{Training efficiency.} Approximate training cost in GPU-days on the indicated hardware. \ourmethod is comparable to Viewset Diffusion and Splatter Image, yet far less expensive than large models (e.g., LRM, LVSM).}
    
    \resizebox{0.8\columnwidth}{!}{
        \begin{tabular}{lcr}
            \toprule
            Method            & GPU Device & Training Time \\
            \midrule
            Viewset Diffusion & L40        & 3 Days        \\
            Splatter Image    & A6000      & 7 Days        \\
            LRM               & A100       & 300 Days      \\
            LVSM              & A100       & 200 Days      \\
            \ourmethod        & L40        & 3 Days        \\
            \bottomrule
        \end{tabular}
    }
    \label{tab:training_efficiency}
    
\end{table}

\section{Conclusion}
We presented \ourmethod, an efficient image-to-3D generation framework that represents scenes with a compact set of learned 3D anchor queries and decodes them into explicit 3D Gaussian primitives. Combined with a 3D position-aware encoder and a transformer-based query-to-Gaussian expansion network trained under rectified flow, \ourmethod enables one-step inference while preserving high-quality, view-consistent novel-view synthesis. Compared to dense 3D initializations or iterative denoising, our sparse set representation exhibits high efficiency and utilization, and thus yields substantial gains in runtime and memory efficiency. Experiments validate effectiveness across quality, input-view bias, and efficiency metrics. Promising directions for future work include query-level controllable editing and scaling to larger unposed in-the-wild captures.

{
    \small
    \bibliographystyle{ieeenat_fullname}
    \bibliography{main}
}

\clearpage
\appendix
\section*{Appendix}

\section{Implementation Details}
\label{sec:impl_details}
In this section, we provide additional implementation details of our method.

\subsection{3D Anchor Queries}

We maintain a bank of $N$ learnable 3D reference points within the model:
\begin{equation*}
    r_i \in \mathbb{R}^3,\qquad i=1,\dots,M,
\end{equation*}
which act as the explicit 3D anchor points.

To obtain the corresponding anchor queries in the hidden space, we encode each reference point with sinusoidal positional embeddings and project it to the query feature dimension using a small MLP (FC--ReLU--FC):
\begin{equation*}
    Q_i = \text{MLP}\big(\text{PE}(r_i)\big) \in \mathbb{R}^d,\qquad i=1,\dots,M,
\end{equation*}
We then feed these queries into the transformer decoder.

\subsection{3D Gaussian Decoding}

After passing each anchor query through the transformer decoder and aggregating image features, we use an MLP-based Gaussian head to regress the 3D Gaussian parameters $(\mu, \Sigma, c, \alpha)$ from the updated query features $Q'_i$. Each query is expanded into $K$ 3D Gaussians (expansion factor).
For $\Sigma$ (scale and rotation), $c$ (color), and $\alpha$ (opacity), we predict them directly from the query features:
\begin{align*}
    \Sigma_{i,j} &= \text{MLP}_{\Sigma}(Q'_i),\quad j=1,\dots,K,\\
    c_{i,j} &= \text{MLP}_{c}(Q'_i),\quad j=1,\dots,K,\\
    \alpha_{i,j} &= \text{MLP}_{\alpha}(Q'_i),\quad j=1,\dots,K.
\end{align*}
where each MLP is a separate small FC--ReLU--FC--ReLU--FC--Activation network, with Sigmoid activation for color, opacity and mean offset, and $exp()$ activation for scale.

For the mean $\mu$, we predict it as an offset from the corresponding reference point $r_i$ of each query:
\begin{equation*}
    \mu_{i,j} = r_i + \text{MLP}_{\mu}(Q'_i),\quad j=1,\dots,K.
\end{equation*}
This offset formulation strengthens the 3D spatial prior induced by the reference points and stabilizes training.

\subsection{Gaussian Parameter Regularization}
To encourage reasonable Gaussian parameter predictions, we apply several regularization terms during training.
First, we hard clip the predicted scale values extracted from $\Sigma$ to lie within a predefined range $[0, s_{max}]$, where $s_{max}$ is a hyperparameter. This prevents the model from predicting excessively large gaussians that could destabilize training.

Second, we apply an $L_2$ regularization loss on the predicted offsets for the means $\mu$ if it exceeds a certain threshold, encouraging the means to stay close to their corresponding reference points:
\begin{equation*}
    \mathcal{L}_{\text{offset}} = \frac{1}{N K} \sum_{i=1}^{N} \sum_{j=1}^{K} \max\big(0, \|\mu_{i,j} - r_i\|_2 - \delta\big)^2,
\end{equation*}
This helps maintain spatial coherence and also strengthens the 3D spatial prior.

\begin{table}[t]
    \centering
    \caption{Default hyperparameters used in our experiments.}
    \vspace{-2mm}
    \label{tab:hyperparams}
    \resizebox{\linewidth}{!}{
        \begin{tabular}{ccl}
            \toprule
            Parameter         & Value     & Description                              \\
            \midrule
            M                 & 512       & Number of anchor queries                 \\
            K                 & 10        & Gaussians per query (expansion factor)   \\
            d                 & 512       & Transformer hidden dimension             \\
            $H, W$            & 128       & Input image resolution (pixels)          \\
            $d_{th}$          & 64        & Depth samples per ray                    \\
            $s_{max}$         & 0.1       & Maximum Gaussian scale                   \\
            $N_{\text{enc}}$  & 6         & Number of encoder layers                 \\
            $N_{\text{dec}}$  & 6         & Number of decoder layers                 \\
            $\delta$          & 0.1       & Threshold for mean-offset regularization \\
            $\lambda_{reg}$   & 0.05      & Parameter regularization weight          \\
            $\lambda_{perc}$  & 0.1       & Perceptual loss weight                   \\
            $\lambda_{inter}$ & 0.1       & Intermediate supervision weight          \\
            $\lambda_{occ}$   & 0.1       & Occupancy loss weight                    \\
            n\_iter           & 300{,}000 & Number of training iterations            \\
            lr                & 2e-5      & Learning rate                            \\
            $\beta_1$         & 0.9       & Adam $\beta_1$                           \\
            $\beta_2$         & 0.99      & Adam $\beta_2$                           \\
            \bottomrule
        \end{tabular}
    }
\end{table}

\begin{figure*}[t]
    \centering
    \begin{subfigure}[t]{0.48\linewidth}
        \centering
        \includegraphics[width=\linewidth, height=1.3\linewidth]{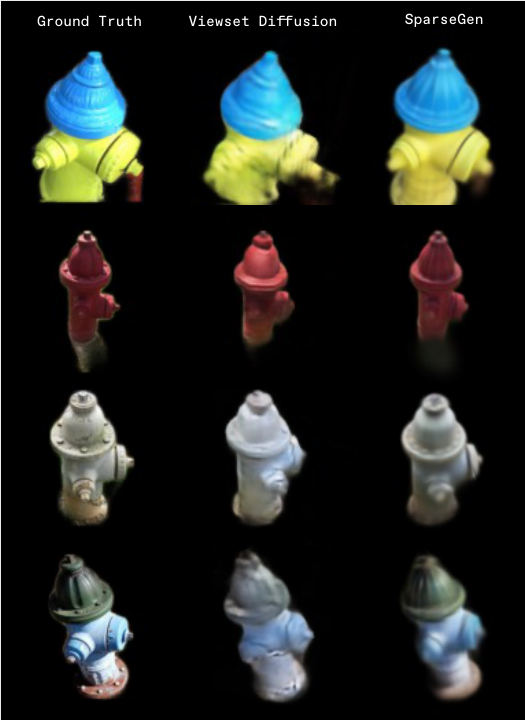}
        \caption{CO3D Hydrant}
        \label{fig:quali_hyd}
    \end{subfigure}
    \hfill
    \begin{subfigure}[t]{0.48\linewidth}
        \centering
        \includegraphics[width=\linewidth, height=1.3\linewidth]{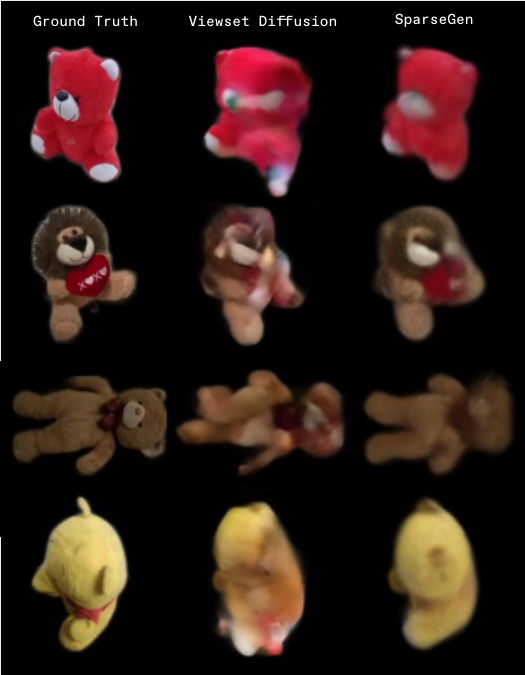}
        \caption{CO3D Teddybear}
        \label{fig:quali_ted}
    \end{subfigure}
    \caption{\textbf{Qualitative results on CO3D subsets.} We show examples under one-view conditioning settings. Compared to other methods, \ourmethod yields better visual quality, with significantly faster speed and smaller representation size.}
    \label{fig:quali_co3d}
\end{figure*}

\subsection{Multi-layer Supervision}
Inspired by literature from autonomous driving and object detection~\cite{hu_PlanningorientedAutonomousDriving_2023}, we apply supervision from multiple decoder layers. Specifically, we extract intermediate query features from each decoder layer and predict Gaussian parameters from them. We then compute the same reconstruction loss on these intermediate predictions as on the final output:
\begin{equation*}
    \mathcal{L}_{\text{inter}} = \frac{1}{N_{\text{dec}} - 1} \sum_{l=1}^{N_{\text{dec}} - 1} \mathcal{L}_{\text{recon}}^{(l)},
\end{equation*}
where $N_{\text{dec}}$ is the total number of decoder layers and $\mathcal{L}_{\text{recon}}^{(l)}$ is the reconstruction loss computed on the predictions from the $l$-th layer. A weight factor $\lambda_{\text{inter}}$ is applied before adding this loss to the total training loss.

\subsection{Hyperparameters}
For reproducibility, we list the default training hyperparameters used in our experiments in \Cref{tab:hyperparams}.

\subsection{Additional Qualitative Results}
\begin{figure*}[p]
    \centering
    \includegraphics[width=\textwidth,height=\textheight,keepaspectratio]{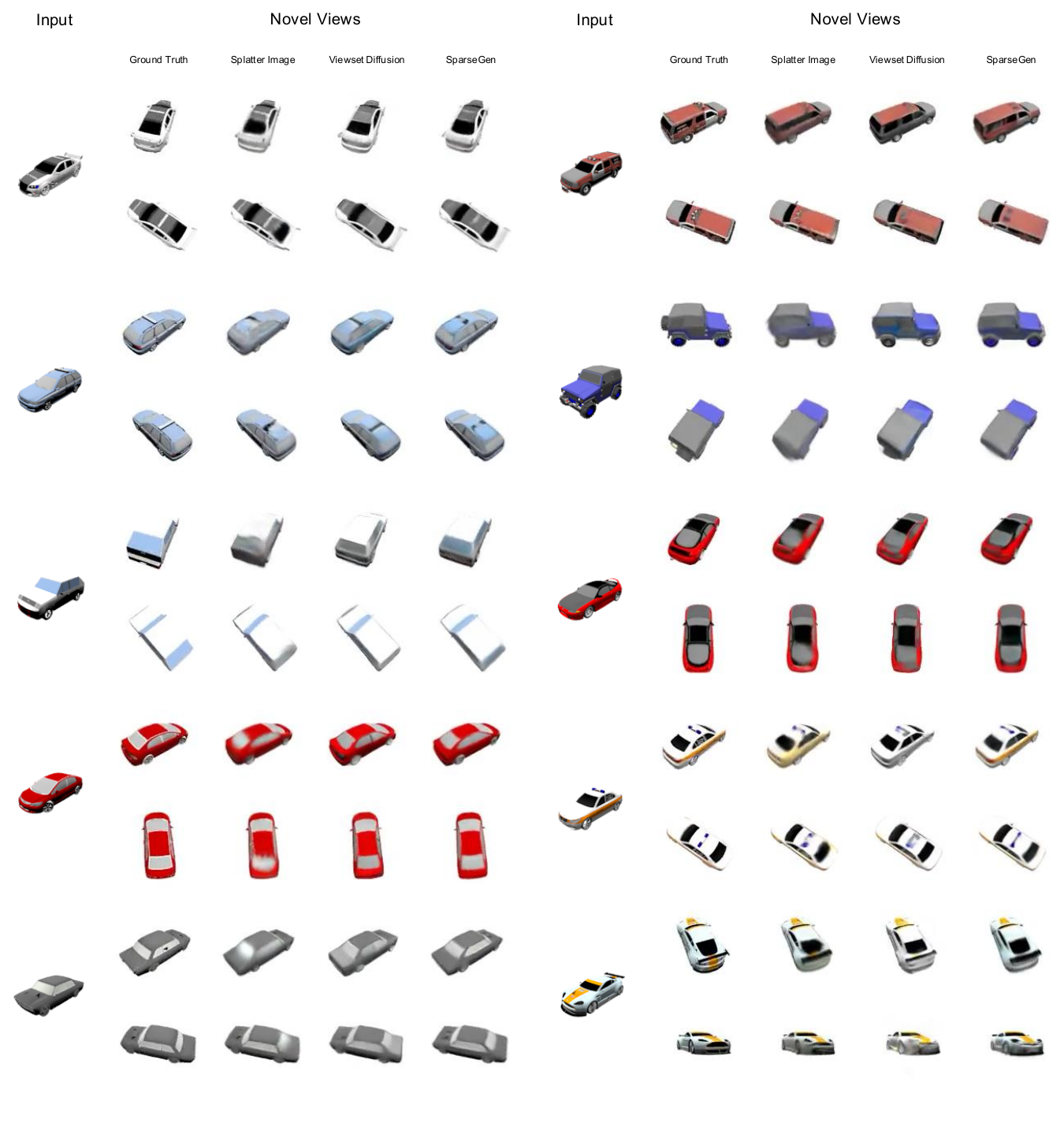}
    \caption{\textbf{Additional qualitative results on the ShapeNet-SRN dataset.} Each group features one input image (left) and two novel view renderings (right) from all tested methods. Generally, deterministic feed-forward methods (i.e., Splatter Image~\cite{szymanowicz_SplatterImageUltraFast_2024}) tend to produce unsatisfactory details on regions not well observed in the input view, while generative methods with iterative diffusion (i.e., Viewset Diffusion~\cite{szymanowicz_ViewsetDiffusion0ImageConditioned_2023}) can synthesize plausible details but sometimes introduce artifacts. Our \ourmethod effectively generates high-quality novel views with ultra-fast speed.}
    \label{fig:shapenet_supp}
\end{figure*}

We provide additional qualitative results on the ShapeNet-SRN dataset in \Cref{fig:shapenet_supp} to complement the main paper. As shown, our \ourmethod effectively generates high-quality novel views with fine details while maintaining ultra-fast inference speed compared to prior methods.

\subsection{Qualitative Results on CO3D}
We include additional qualitative comparisons on CO3D (Hydrant and Teddybear) in \Cref{fig:quali_hyd,fig:quali_ted}. These figures complement the quantitative evaluation in \Cref{tab:co3d_subsets} in the main paper.

\section{Additional Results and Discussions}
\label{sec:add_res}

\subsection{Generalization to In-the-wild Objects}

To evaluate the generalization capability of our method to diverse in-the-wild objects, we conduct additional experiments by training on the renderings of the Objaverse dataset~\cite{deitke_ObjaverseUniverseAnnotated_2023}, and testing on the Google Scanned Objects (GSO) dataset~\cite{downs_GoogleScannedObjects_2022}. All experiments are performed at a resolution of $128\times128$.

As shown in \Cref{tab:objaverse}, \ourmethod achieves competitive performance compared to prior methods, attaining the highest PSNR while maintaining favorable perceptual quality (LPIPS and SSIM), even with a much faster inference speed and smaller representation size (0.033s and 560KB, with other methods same as the main paper). These results are still preliminary, and we expect \ourmethod to further improve and scale to higher fidelity with additional training compute and larger-scale data.

\begin{table}[t]
\centering
\caption{\textbf{Quantitative evaluation on the GSO dataset.}
Best results in \textbf{bold}, second best \underline{underlined}. Our method achieves the highest PSNR while maintaining competitive perceptual quality.}
\vspace{-1mm}
\label{tab:objaverse}
\resizebox{0.8\linewidth}{!}{
    \begin{tabular}{l|ccc}
    \toprule
    Method                    & PSNR $\uparrow$   & LPIPS $\downarrow$ & SSIM $\uparrow$  \\
    \midrule
    OpenLRM                   & 14.526 & 0.199 & 0.741 \\
    Splatter Image            & \underline{21.065} & \textbf{0.111} & \textbf{0.878} \\
    \ourmethod & \textbf{21.427} & \underline{0.160} & \underline{0.850} \\
    \bottomrule
    \end{tabular}
}
\end{table}

\end{document}